\RequirePackage{amsmath}
\documentclass[runningheads]{llncs}
\usepackage[T1]{fontenc}
\usepackage{graphicx}
\usepackage{float}
\usepackage{hyperref}
\usepackage{color}

\begin{document}
%
\title{An End-to-End Approach for Child Reading Assessment in the Xhosa Language}
\titlerunning{An End-to-End Approach for Child Reading Assessment in Xhosa}
%
\author{Sergio Chevtchenko\inst{1,2} \and
Nikhil Navas\inst{1,2} \and
Rafaella Vale\inst{4} \and
Franco Ubaudi\inst{5} \and
Sipumelele Lucwaba\inst{6} \and
Cally Ardington\inst{3} \and
Soheil Afshar\inst{7} \and
Mark Antoniou\inst{2} \and
Saeed Afshar\inst{1,2}}

\authorrunning{S. Chevtchenko et al.}
\institute{International Centre for Neuromorphic Systems, Western Sydney University, Australia
\email{}\\ 
\url{s.chevtchenko@westernsyndey.edu.au} \\
\and
The MARCS Institute for Brain, Behaviour and Development, Western Sydney University, Australia
\and
University of Cape Town, South Africa
\and
Universidade Federal de Pernambuco, Centro de Informatica, Recife, PE, Brazil
\and
The School of Computer, Data and Mathematical Sciences, Western Sydney University, Australia\\
\and
University of Pretoria, Gauteng, South Africa
\and
Macquarie University, Australia\\
}
\maketitle              
\begin{abstract}

Child literacy is a strong predictor of life outcomes at the subsequent stages of an individual's life. This points to a need for targeted interventions in vulnerable low and middle income populations to help bridge the gap between literacy levels in these regions and high income ones. In this effort, reading assessments provide an important tool to measure the effectiveness of these programs and AI can be a reliable and economical tool to support educators with this task. Developing accurate automatic reading assessment systems for child speech in low-resource languages poses significant challenges due to limited data and the unique acoustic properties of children's voices. This study focuses on Xhosa, a language spoken in South Africa, to advance child speech recognition capabilities. We present a novel dataset composed of child speech samples in Xhosa. The dataset is available upon request and contains ten words and letters, which are part of the Early Grade Reading Assessment (EGRA) system. Each recording is labeled with an online and cost-effective approach by multiple markers and a subsample is validated by an independent EGRA reviewer. This dataset is evaluated with three fine-tuned state-of-the-art end-to-end models: wav2vec 2.0, HuBERT, and Whisper. The results indicate that the performance of these models can be significantly influenced by the amount and balancing of the available training data, which is fundamental for cost-effective large dataset collection. Furthermore, our experiments indicate that the wav2vec 2.0 performance is improved by training on multiple classes at a time, even when the number of available samples is constrained. 

\end{abstract}
\section{Introduction}
\label{sec_Intro}

Reading assessment in the early grades is essential for pinpointing the difficulties children encounter while learning to read. By identifying these challenges, educators can design targeted interventions that accelerate literacy development and foster inclusion. Artificial intelligence offers a practical way to scale such assessments, providing rapid insights into children's oral-reading skills and the obstacles revealed by pronunciation tasks, even when data and resources are limited.

The motivation for this research is a pragmatic one. Good education is a crucial component for facilitating a favorable future for children. While assessing speech fluency and reading skills is a valid way to inform about the advancement of literacy acquisition in children~\cite{Dubeck2015}, a simple approach to evaluating their aptitude to understand text in early years is to analyze their ability to correctly pronounce words~\cite{Namasivayam2020}, a key focus of this research.
A poor ability to pronounce words is indicative of possible issues with hearing, understanding grammar, or other difficulties such as dyslexia~\cite{Sobti2024}.
As a result, uncovering these issues quickly can promote focused guidance from an early age that greatly improves a child’s educational outcome~\cite{kim2022,Sobti2024}.

To tackle this, the use of automatic speech recognition (ASR) can greatly assist in finding reading issues that manifest themselves in pronunciation difficulties. ASR can help automate thorough and accurate assessment of a child’s ability and the determination of progress as educators seek to address problems. However, the effectiveness of ASR can be greatly impacted in the context of low-resource languages (LRL)~\cite{magueresse2020low,ranathunga2023low}, that is, languages with limited availability of resources, for the effective automated construction of models needed for assessing children’s speech. Hence this research also focuses on languages that are characterized as LRL, an example of which is the South African language Xhosa, which also provides phonetic complexity challenges to this research~\cite{mesham2021low}.

This research investigates the use of state-of-the-art (SOTA) speech classification models that convert the audio of child speech into the phonetics responsible for the language of choice. The South African language Xhosa is used as the target language. Our investigation is principally focused on testing a cost-effective data-labeling method and understanding the impact of choices in model parameters, training set sizes, and model building strategies. The main goal of this research is the development of a solution that fulfills the following criteria: accurate assessment of a child’s reading ability; coping with restrictions in terms of the amount of available training data; potential integration with the Early Grade Reading Assessment (EGRA) program~\cite{Dubeck2015}, and providing a solution architecture that is easy to reuse for other speaking languages. 

Accurately assessing a child’s ability to pronounce words is time-consuming and prone to error, since it can be impacted by the ability and consistency of individual human assessors. As such, automation provides a way for avoiding these issues, while also reducing needed time and effort. Additionally, AI-powered assessments provide a means for consistently tracking a child’s progress, while reducing the impact of confounding variables such as human assessor subjectivity and enforcing the use of an identical assessment across time and individual children.

The next problem to address is regarding LRLs.
Many speaking languages are characterized as LRLs because of a lack of large text corpora or other linguistic resources that are suitable for training ASR solutions.
This presents a serious barrier since it impacts solution effectiveness and greatly increases the needed effort for developing ASR solutions. Novel approaches are needed for facilitating the development of accurate models, coping with limited resources, and managing variability of human expertise across limited staff. Given these challenges, the use of self-supervised SOTA speech recognition architectures offers a promising solution by reducing the reliance on extensive labeled datasets and enabling a more detailed focus on nuanced parts of speech.

The present work makes the following contributions to automated reading assessment for LRL. First, it introduces a new, high-quality dataset of Xhosa child reading samples, specifically designed for early grade reading assessment. This dataset has been made available to the community~\cite{chevtchenko2025dataset}. Second, it details and validates a cost-effective online method for data labeling and verification, demonstrating its suitability for resource-constrained settings. Third, it provides a comprehensive evaluation of three fine-tuned SOTA speech classification architectures on this dataset, highlighting how variations in training set size and class balancing can significantly impact performance.





\section{Related Work}
\label{sec_related_work}

Among self-supervised models, wav2vec has emerged as a state-of-the-art architecture for ASR, using self-supervised learning to extract meaningful features from raw audio and requiring minimal labeled data for fine-tuning~\cite{baevski2020wav2vec}. Doumbouya et al.~\cite{doumbouya2021radio} adapted wav2vec to the West African Radio Corpus, demonstrating sensitivity to features of African languages and effective identification in noisy multilingual settings. Similarly, our work involves fine-tuning a pre-trained wav2vec 2.0 model, augmented with a classifier multilayer perceptron to classify audio recordings of early grade children learning the Xhosa language as correct or incorrect pronunciations.

Wav2vec has also been effective in detecting pronunciation errors, offering potential in language learning and pronunciation assessment~\cite{peng2023md}. Shekar et al.~\cite{shekar2023gop} used wav2vec to analyze speech recordings of non-native speakers, detecting mispronunciations and capturing fine-grained phonetic details for phoneme-level error diagnosis. Recent research by Jain et al.~\cite{jain2023child} examined the use of wav2vec models in pediatric learning assessment, a challenging domain due to varied speech patterns in children. Abaskohi et al.~\cite{abaskohi2023persian} further developed and adapted wav2vec with a preprocessing technique called random frequency pitch, useful for capturing the frequency characteristics of children's speech. Applied to Persian children's speech and reading recordings, the model significantly reduced word error rates compared to baselines, confirming its effectiveness in pediatric assessment tasks. Obiang et al.~\cite{obiang2024yoruba} fine-tuned wav2vec for Yoruba, a low-resourced tonal language, effectively capturing complex pitch and pronunciation variations superior to traditional methods. Similarly, Okwugbe et al.~\cite{okwugbe2024fonigbo} developed a deep learning model combining rCNNs and BiLSTMs with CTC for speech recognition of the Fon and Igbo languages, addressing challenges due to data scarcity.

Other works have tackled the application of ASR in assessing and improving reading skills, particularly in children. Sabu and Rao~\cite{sabu2018automatic} sought to evaluate word reading accuracy and identify prosodic elements like phrase breaks and emphasized words in children studying English as a second language. The complexities of the prosodic events were reflected in the lower precision estimation shown by the system, revealing the importance of an accurate feature extraction. Shivakumar and Georgiou~\cite{shivakumar2020transfer} used transfer learning from adult ASR models as a way to circumvent the lack of extensive availability of children speech data. Strategies to adapt to varying amounts of data and children's ages were also recommended. Bai et al.~\cite{bai20b_interspeech} explored the evaluation of decoding skills in Dutch children to provide feedback meant to impact reading accuracy and speed. The study was further expanded in~\cite{bai2021automatic} to also highlight the most problematic words encountered by the students.
Bachiri et al.~\cite{bachiri2024integrating} proposed an interactive learning system as a Moodle plugin to aid in child reading assessment in underserved educational settings. The authors make a case for the integration of AI-based speech recognition to enhance literacy development due to the positive results observed. In an attempt to investigate how AI performs in assessing students reading fluency and level compared to human experts, Y{\i}ld{\i}z et al.~\cite{yildiz2025can} found that AI offered reliable predictions that could contribute to lowering costs and improving efficiency.

The recent publication by Henkel et al.~\cite{henkel2025literacy} has a significant overlap with the intended goals of the current work. The automation of reading assessment for students in a resource-constrained context is investigated to compare the potential of Whisper V2 and wav2vec 2.0 models against human raters. The models produce transcriptions of audio recordings of Ghanaian students reading to provide an estimate of the students' oral reading fluency. The fluency scores given by Whisper V2 were found to be highly correlated with assessments by human experts, even without fine-tuning. It is important to note that the way these models were used to score reading fluency in students would not be suitable for the goal of the current work, since the lack of fine-tuning may result in overcorrections that ignore mispronunciations or in the misinterpretation of accents, as demonstrated in Section~\ref{ssec_asr_com_mms}.
As the present work aims to evaluate the pronunciation of words and letters by children in the Xhosa language, fine-tuning of the models is necessary to classify samples as correct or incorrect according to the target language.

The above studies underscore the effectiveness of speech-to-text models for low-resource languages and child speech recognition. To the best of our knowledge, this is the first work to evaluate end-to-end models on words and letters for a child reading assessment problem.

\section{Methodology}
\label{sec_Methodology}



\subsection{Dataset Description}
\label{ssec_dataset}

The dataset was collected in 2024 as part of the Early Grade Reading Assessment (EGRA), conducted in grades 1 to 4 at schools in South Africa, and is available online~\cite{chevtchenko2025dataset}. This study focuses on the following ten letters and words in the Xhosa language: \textit{d}, \textit{v}, \textit{n}, \textit{ewe}, \textit{hayi}, \textit{hl}, \textit{inja}, \textit{kude}, \textit{molo} and \textit{ng}. A question is presented to the student, asking to pronounce one of these words or letters, followed by a recording. The recording stops when the student presses a button for the next question. The average duration of the recordings is around 4 seconds, with added variability mostly due to multiple attempts at pronouncing the same question. 

\begin{figure}[ht!]
    \centering
    \includegraphics[width=\linewidth]{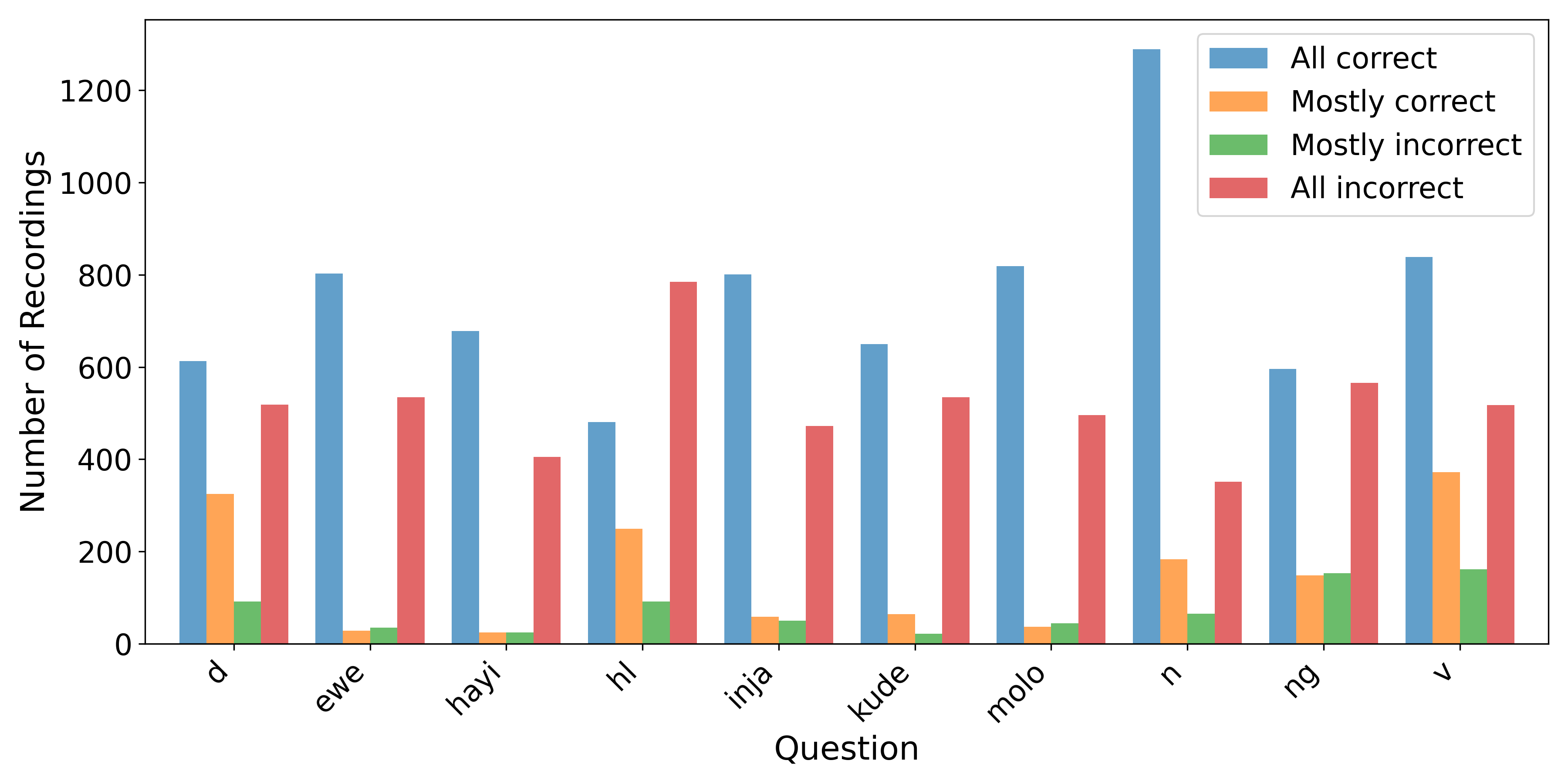}
    \caption{Distribution of All correct, Mostly correct, Mostly incorrect, and All incorrect samples for each word and letter in the dataset.}
    \label{fig_all_vs_mostly_correct_by_word}
\end{figure}

A total of 14,971 recordings of the EGRA collection with over 19 hours in recording time collected over 8 months are made available upon request. Each recording is labeled as correct or incorrect by three fluent Xhosa speakers, resulting in a dataset of 44,913 labeled entries. Since some recordings received conflicting labels, a validation experiment was conducted, as described in Section \ref{ssec_labeling}. Figure~\ref{fig_all_vs_mostly_correct_by_word} illustrates the distribution of these labels by question, showing that certain items can be more challenging to classify. For example, most children correctly pronounced the letter \textit{n}, whereas a considerable number struggled with the consonant \textit{hl}. To enable cost-effective and efficient data collection and labeling, custom mobile and server applications were developed as part of the project. A more detailed description of these tools will be provided in a forthcoming publication. 

\begin{figure}[ht!]
    \centering
    \includegraphics[width=0.8\linewidth]{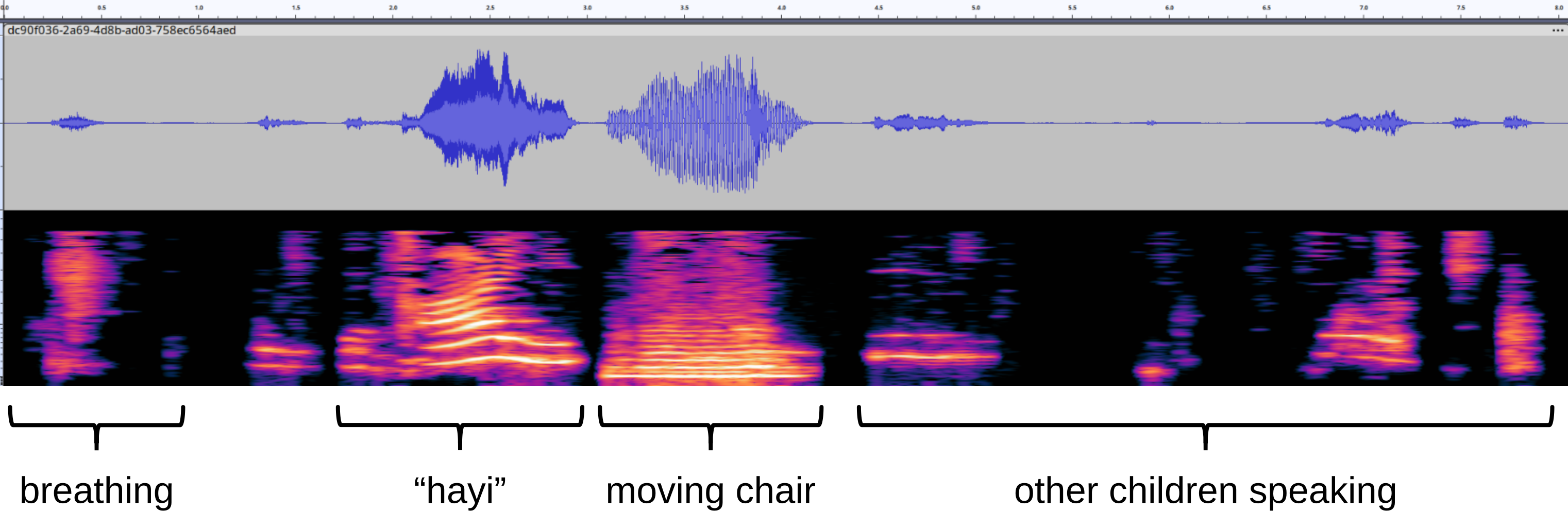}
    \caption{A spectrogram of an eight-second recording containing a correct pronunciation of the word ``hayi'' and other ambient sounds.}
    \label{fig_hayi_spectrogram_commented}
\end{figure}

Although each recording is intended to capture a single spoken word, even those pronounced correctly often include additional artifacts, such as classroom noise or multiple attempts at pronouncing the word on the app's screen.
This variability in recording size and quality makes the classification problem more complicated and is further illustrated in Figure~\ref{fig_hayi_spectrogram_commented}.

\subsection{Data Labeling and Validation}
\label{ssec_labeling}

\begin{figure}[ht!]
    \centering
    \includegraphics[width=\linewidth]{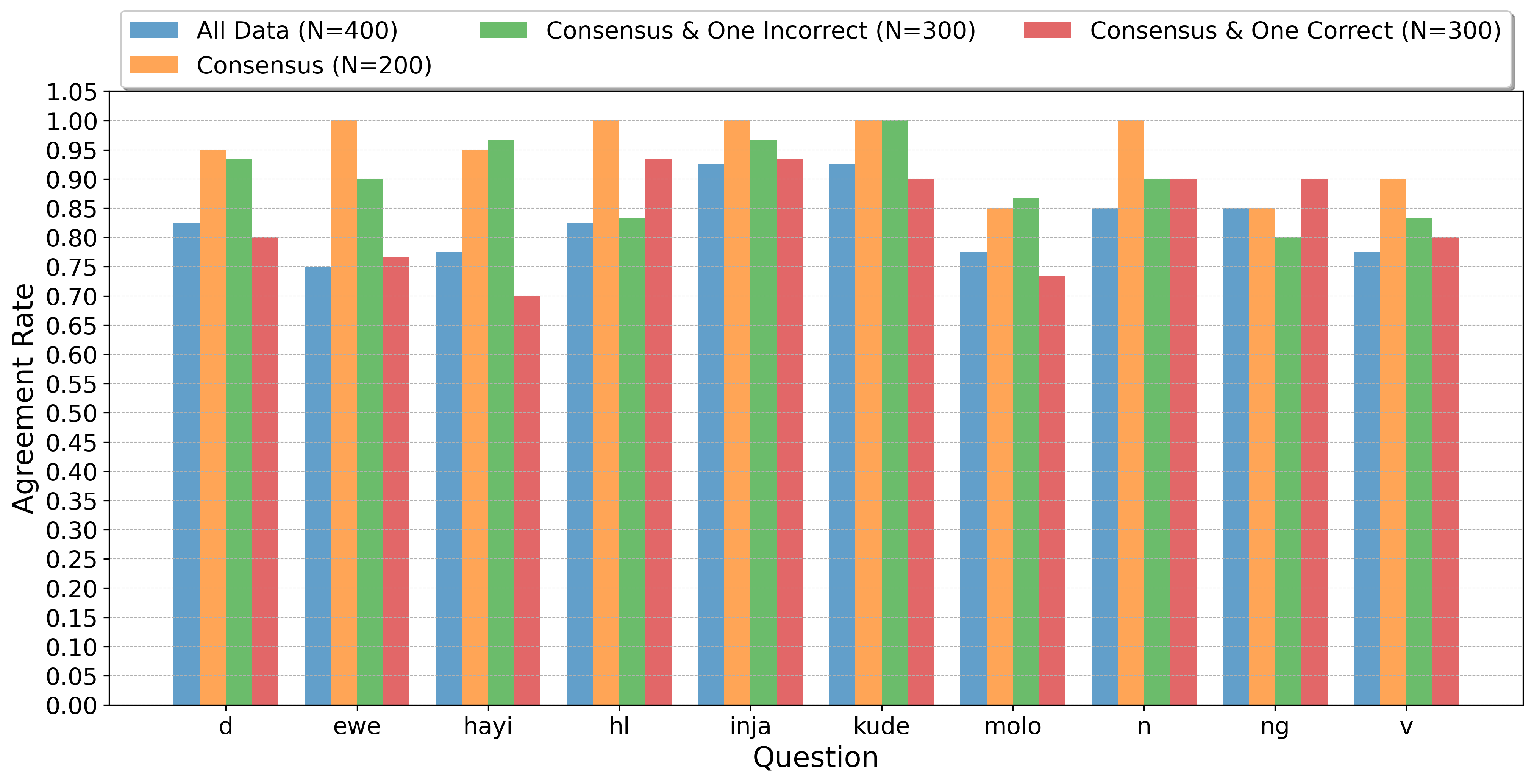}
    \caption{Agreement rate between an expert and the reviewers, considering different validation criteria.}
    \label{fig_agreement_rate}
\end{figure}

 Selecting appropriate validation criteria for training data is another significant challenge. To ensure high-quality recordings, each sample is labeled by three independent markers, and only those with consensus are included in training. To validate this method, 400 recordings were randomly selected from ten words in the dataset, with each word contributing 40 samples. These were evenly distributed across four marking scenarios: (a) all three markers labeled it as correct, (b) all three labeled it as incorrect, (c) one marked it incorrect while the other two marked it correct, and (d) one marked it correct while the other two marked it incorrect. An expert in traditional EGRA marking for Xhosa language then assessed these samples, and Table~\ref{tab_agreement_rates} presents the agreement rate between the expert and the original markers. An agreement rate per question is also presented in Figure \ref{fig_agreement_rate}. Thus, consensus-based approach retains approximately 85\% of the original dataset, while maximizing data reliability. Consequently, while the provided dataset includes all 14,971 recordings, the experimental results in Section \ref{sec_results} are based only on the 12,747 consensus-marked recordings. 

\begin{table}
\centering
\caption{Agreement rates and number of recordings for each condition.}
\label{tab_agreement_rates}
\begin{tabular}{|l|l|l|}
\hline
Condition                  & Agreement rate & Total \# of recordings \\
\hline
All data                   & 71.25\%        & 14,971               \\
Consensus                  & 85.00\%        & 12,747 (85.1\%)      \\
Consensus \& one incorrect & 80.00\%        & 14,235 (95.0\%)      \\
Consensus \& one correct   & 75.00\%        & 13,483 (90.0\%)      \\
\hline
\end{tabular}
\end{table}

\subsection{Fine-tuned Classification Architectures}
\label{ssec_algorithms}

In this work, we consider three SOTA speech recognition architectures: wav2vec 2.0~\cite{baevski2020wav2vec}, HuBERT~\cite{hsu2021hubert} and Whisper~\cite{radford2023robust}. In order to select baseline models, we conducted preliminary experiments with the pre-trained versions of these architectures and varying hyperparameters such as learning rate and batch size. While this is not meant to be an exhaustive optimization, this step is necessary to provide a good baseline for the presented dataset. The pre-trained models are fine-tuned with a learning rate of $3 \times 10^{-5}$ over 1000 steps with a batch size of four and gradient accumulation of over two steps to achieve an effective batch size of eight.

This study used a snapshot of the wav2vec\footnotemark[1]{}\footnotetext[1]{https://huggingface.co/facebook/wav2vec2-xlsr-53-espeak-cv-ft} 2.0 model which has been pre-trained on the Multilingual LibriSpeech\footnotemark[2]{}\footnotetext[2]{https://www.openslr.org/94/}, CommonVoice\footnotemark[3]{}\footnotetext[3]{https://commonvoice.mozilla.org/en/languages}, and BABEL\footnotemark[4]{}\footnotetext[4]{https://catalog.ldc.upenn.edu/byyear} datasets. This extensive training allows for cross-lingual representation learning via a self-supervised approach that extracts meaningful features from raw audio data. Additionally, we employed the multilingual 
mHuBERT~147\footnotemark[5]{}\footnotetext[5]{https://huggingface.co/utter-project/mHuBERT-147} model, which was trained on 90,430~hours of openly licensed audio from 147~languages across 16~independent datasets. Finally, we selected a distilled version of Whisper\footnotemark[6]{}\footnotetext[6]{https://huggingface.co/distil-whisper/distil-medium.en} from OpenAI, an end-to-end Transformer-based architecture for sequence-to-sequence tasks. 

\subsection{Experimental Evaluation and Metrics}
\label{ssec_evaluation}





A test set of 50 positive and 50 negative samples is randomly selected from the dataset for each of the ten questions, resulting in a test set of 1000 recordings. In order to produce more reliable results, we repeat this partitioning five times and evaluate each model's configuration on these five independent runs. Thus, most of the results presented in Section \ref{sec_results} are an average of 50 experiments: 5 randomly selected test sets and 10 questions from the dataset.

One of the main objectives of this work is to assess how varying amounts of training data affect the baseline models. As discussed in Section \ref{ssec_dataset}, certain questions often yield unbalanced samples for correct (positive) and incorrect (negative) pronunciations. To address this, we train the baseline models on 50, 100, 200, and 300 randomly selected positive and negative samples for each question, resulting in 16 distinct training configurations. Each model may be trained on a single question (as a binary classifier) or on multiple questions simultaneously, applying the same sampling strategy per question. 

As each fine-tuned classifier is evaluated on a balanced set of 50 positive and 50 negative samples per question, we define the confusion matrix categories as follows:
\begin{itemize}
    \item \textbf{True positive (TP)}: When both the ground truth label and the prediction correspond to the question.
    \item \textbf{True negative (TN)}: When neither the ground truth label nor the prediction correspond to the question.
    \item \textbf{False negative (FN)}: When the ground truth label corresponds to the question but the prediction does not.
    \item \textbf{False positive (FP)}:  When the ground truth label does not correspond to the question but the prediction does.
\end{itemize}

We then define the false positive rate (FPR) as $FPR = \frac{FP}{FP + TN}$, the false negative rate (FNR) as $FNR = \frac{FN}{TP + FN}$, and the diagnostic efficiency (DE) as $DE = \frac{TP + TN}{TP + TN + FP + FN}$.
 
\section{Results and Discussion}
\label{sec_results}

\begin{figure}[ht!]
    \centering
    \includegraphics[width=0.9\linewidth]{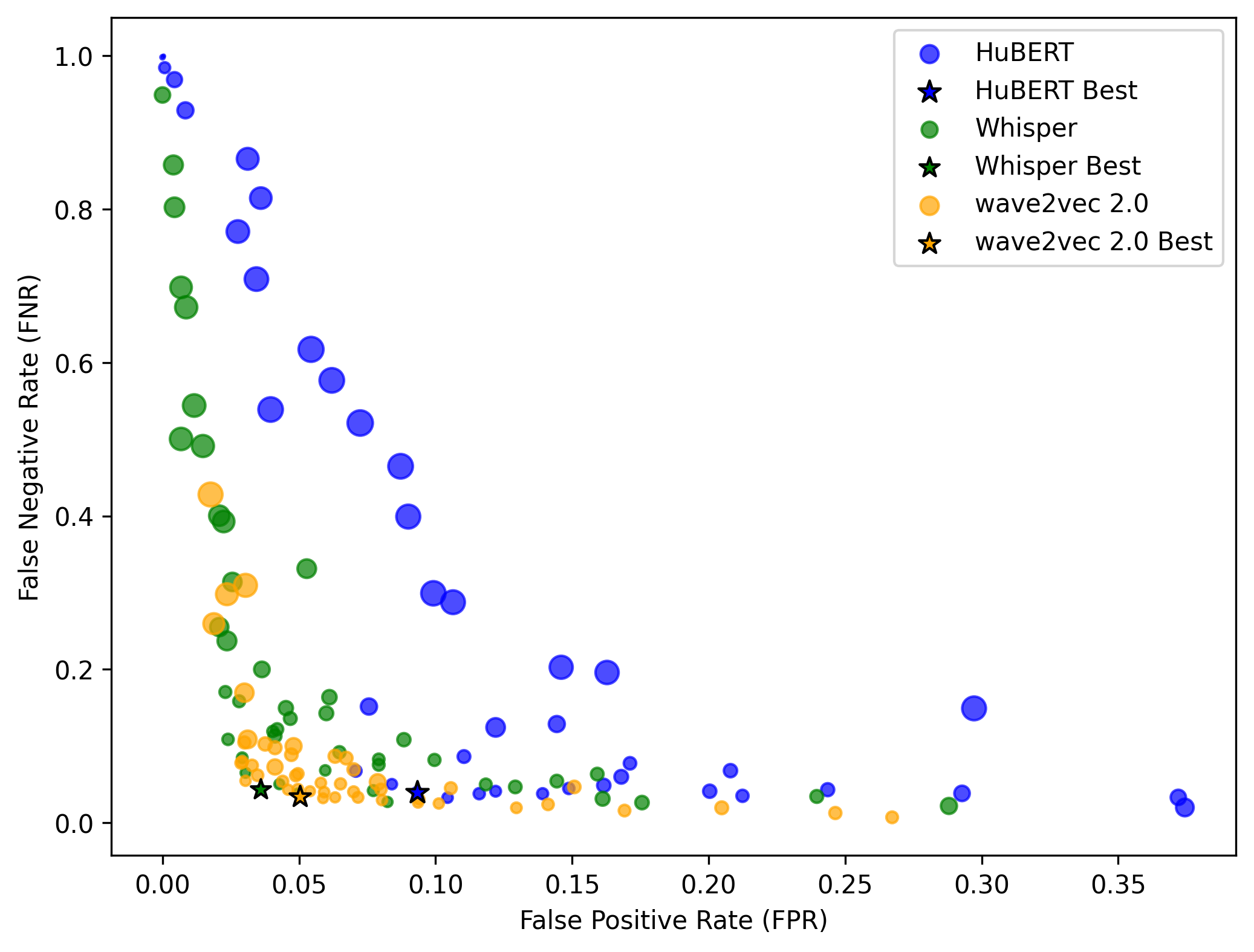}
    \caption{Distribution of FPR and FNR for each model, averaged across ten questions. The dot size is proportional to the standard deviation over 50 samples. The highlighted results are the best ones in terms of the diagnostic efficiency.}
    \label{fig_fpr_fnr_scatter}
\end{figure}

Figure~\ref{fig_fpr_fnr_scatter} provides an overview of the false positive and false negative rates obtained on the test set by the three baseline architectures. As described in Section~\ref{ssec_evaluation}, the models are trained on a variable number of samples and questions at a time, while evaluated on a fixed test subset. The presented data is an average across 10 questions and 5 independent runs, making a total of 50 samples. The results indicate that changes in the training data can have a significant impact on the performance of the models and both wav2vec 2.0 and Whisper generally achieve more favorable FNR-FPR trade-offs than the HuBERT architecture. 

Additionally, Table~\ref{tab_computational_cost} compares the computational requirements of these architectures. While HuBERT is significantly more efficient in terms of training and inference, the wav2vec 2.0 model requires less computational resources for training than Whisper. 

\begin{table}[!ht]
\footnotesize
\caption{%
Comparison of training (1000 steps) and inference time (single recording) on a Google Colab L4 instance, averaged over 10 runs for each model.
}
\centering
\begin{tabular}{lcc}
\hline
\textbf{Model} & \textbf{Training time (s)} & \textbf{Inference time (s)} \\
\hline
HuBERT      & 343  & 0.44 \\
wav2vec 2.0 & 1332 & 1.70 \\
Whisper     & 2511 & 1.54 \\
\hline
\end{tabular}
\label{tab_computational_cost}
\end{table}

\begin{figure}[ht!]
    \centering
    \includegraphics[width=0.6\linewidth]{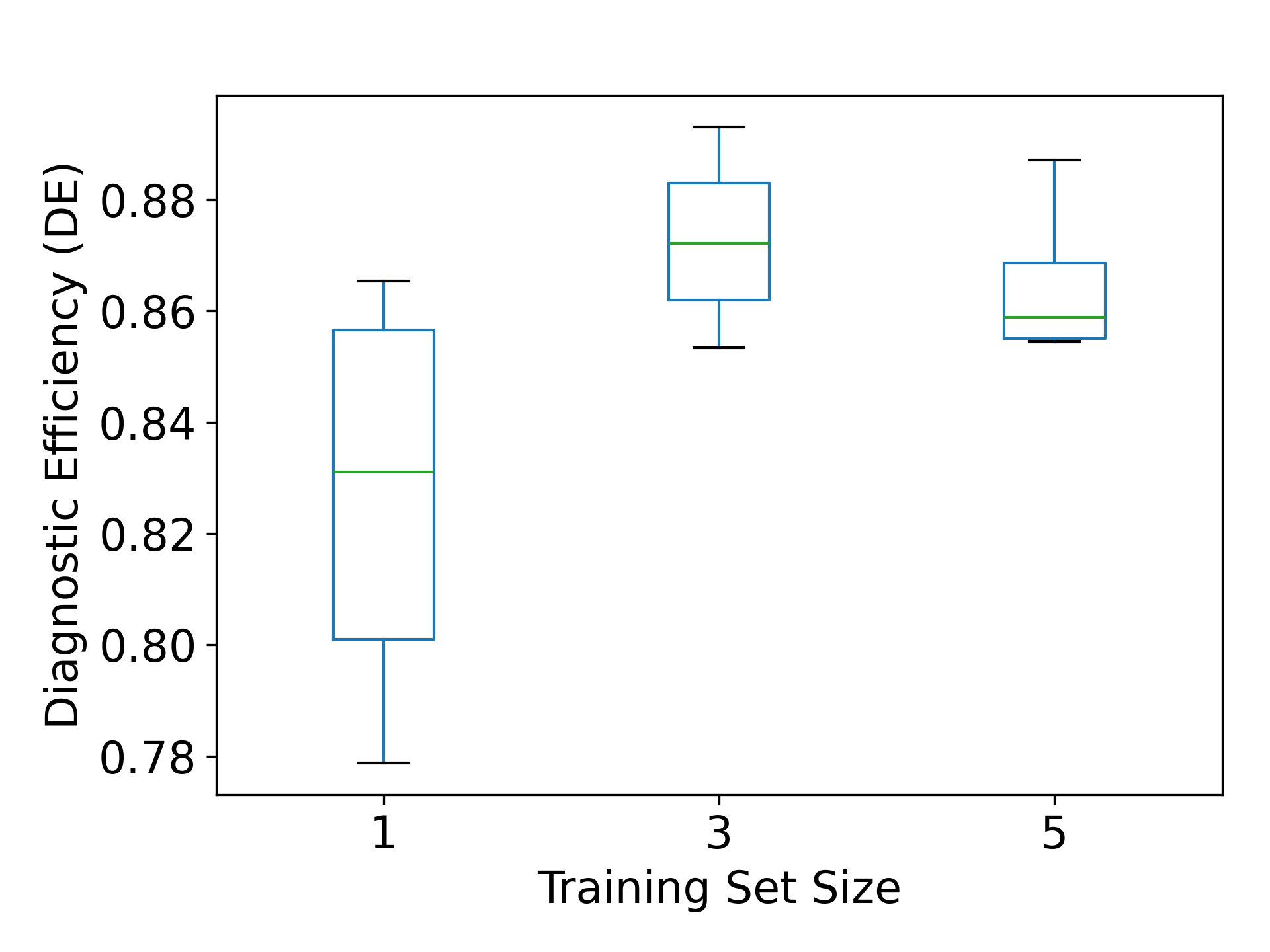}
    \caption{The effect of training set size on the wave2vec 2.0 performance under limited data conditions.}
    \label{fig_wave2vec_de_boxplot_limited}
\end{figure}

The best results obtained from each model are also presented in more detail in Table \ref{tab_top5}, considering top five configurations in terms of the diagnostic efficiency. While wav2vec 2.0 and Whisper provide mostly equivalent results in terms of DE, it should be noted that no single configuration was able to obtain the overall best performance in terms of both FPR and FNR and that wav2vec 2.0 model benefits from being trained on multiple questions at a time. This effect is further illustrated in Figure \ref{fig_wave2vec_de_boxplot_limited}, which shows how training set size influences diagnostic efficiency under a limited-data scenario (i.e., when correct and incorrect samples are restricted to 50 or 100). 

\begin{table}[ht!]
\small
\centering
\caption{Top-five results per model in terms of Diagnostic Efficiency (DE). Highlighted entries are not significantly different (p>0.05) from the best value in each column. Set column indicates the number of questions used to train the model at a time.}
\label{tab_top5}
\begin{tabular}{lcccccc}
\hline
 Model   &   Set &   Correct &   Incorrect & DE \% (±std)          & FPR \% (±std)        & FNR \% (±std)        \\
\hline
 HuBERT  &       1 &       200 &         300 & 87.04 (5.41)          & 9.32 (4.79)          & \textbf{4.00 (3.64)} \\
 HuBERT  &       1 &       100 &         200 & 86.99 (5.55)          & 8.40 (4.06)          & 5.04 (4.26)          \\
 HuBERT  &       1 &       100 &         300 & 86.66 (7.11)          & 7.08 (5.08)          & 6.76 (5.39)          \\
 HuBERT  &       1 &       300 &         300 & 86.61 (5.51)          & 10.44 (4.97)         & \textbf{3.28 (3.42)} \\
 HuBERT  &       1 &       200 &         200 & 85.08 (7.35)          & 11.60 (6.31)         & \textbf{3.80 (3.57)} \\
 W2v 2.0 &       3 &       300 &         200 & \textbf{91.70 (5.28)} & 5.04 (3.50)          & \textbf{3.44 (4.10)} \\
 W2v 2.0 &       3 &       200 &         300 & \textbf{91.66 (6.02)} & \textbf{3.04 (2.56)} & 5.48 (5.46)          \\
 W2v 2.0 &       5 &       300 &         200 & \textbf{91.30 (5.65)} & 4.60 (3.34)          & 4.32 (4.56)          \\
 W2v 2.0 &       5 &       300 &         100 & \textbf{91.10 (4.88)} & 5.88 (3.44)          & \textbf{3.20 (3.92)} \\
 W2v 2.0 &       3 &       200 &         200 & \textbf{90.86 (5.62)} & 4.96 (3.71)          & 4.40 (4.59)          \\
 Whisper &       1 &       300 &         300 & \textbf{92.22 (4.99)} & \textbf{3.60 (3.38)} & 4.36 (3.40)          \\
 Whisper &       1 &       300 &         200 & \textbf{91.41 (5.22)} & 4.96 (3.94)          & \textbf{3.84 (3.28)} \\
 Whisper &       1 &       200 &         200 & \textbf{90.89 (5.19)} & 4.28 (3.33)          & 5.04 (4.45)          \\
 Whisper &       1 &       200 &         300 & \textbf{90.65 (5.40)} & \textbf{3.04 (2.84)} & 6.52 (4.67)          \\
 Whisper &       1 &       300 &         100 & 89.28 (5.75)          & 8.24 (4.96)          & \textbf{2.72 (2.93)} \\
\hline
\end{tabular}
\end{table}

\begin{figure}[ht!]
    \centering
    \includegraphics[width=0.99\linewidth]{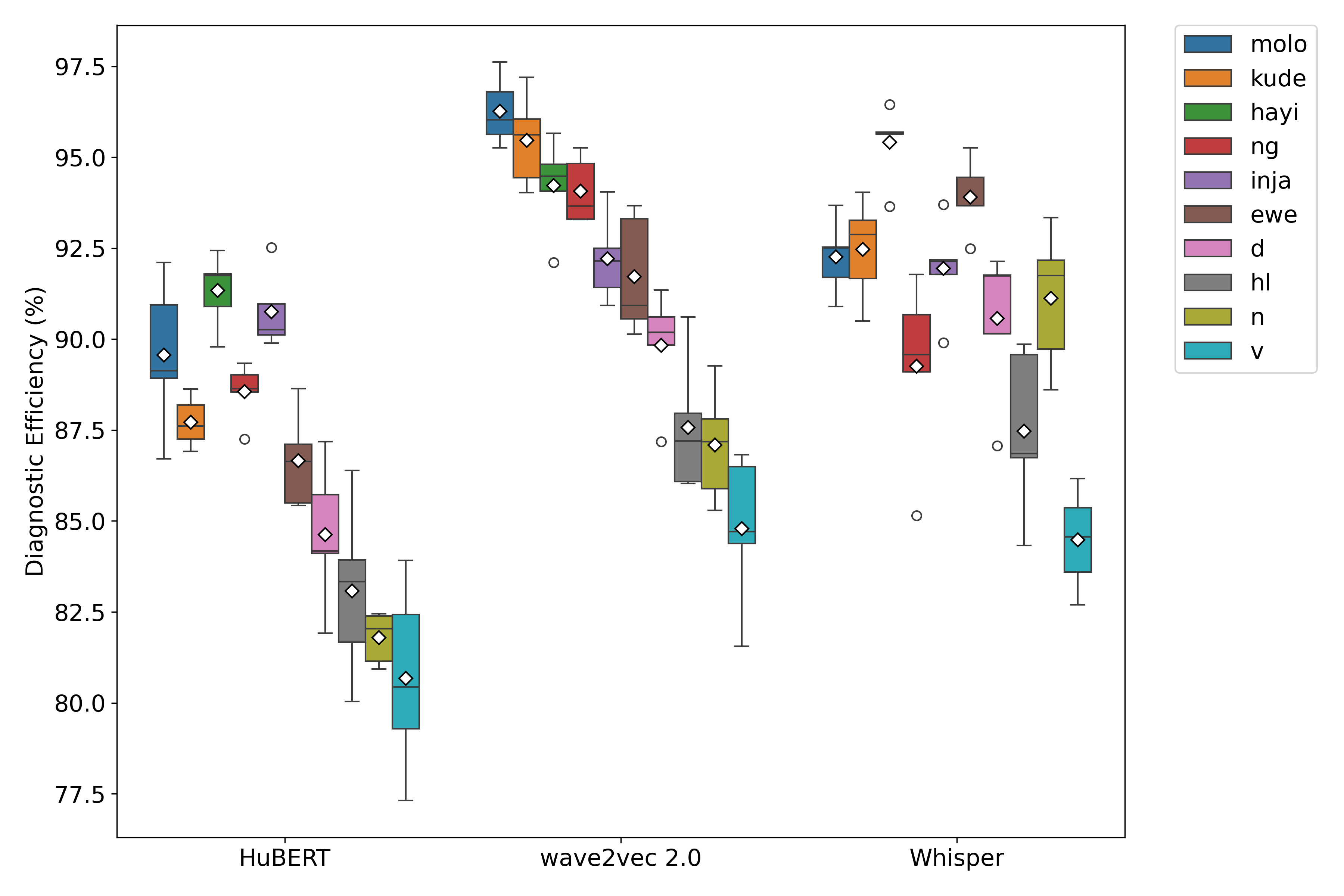}
    \caption{Per-question average performance of the top five models from Table \ref{tab_top5}.}
    \label{fig_top_5_de_boxplot_grouped_by_model_question}
\end{figure}

A per-question breakdown of the top five models is presented in Figure \ref{fig_top_5_de_boxplot_grouped_by_model_question}. This suggests that while an average diagnostic efficiency of around 91\% is possible, the presented dataset is especially challenging for short sounds. 

\subsection{Comparison with an ASR Model}
\label{ssec_asr_com_mms}

In order to highlight the advantages of the classification approach proposed in this work, we also evaluate a recently released checkpoint of the wav2vec 2.0 architecture, fine-tuned on 1,162 languages, including Xhosa, using the Massive Multilingual Speech (MMS-1B) framework \cite{pratap2023mms}.
For testing, we supplied the correct pronunciations of each question in our dataset as input to MMS-1B in speech-to-text mode, paired with a Xhosa tokenizer.
Overall, only 6.91\% of these transcriptions precisely matched the expected word or letter.
Table~\ref{tab_mms_18_results} provides a breakdown of the model’s performance for each question, indicating that while this approach performs better on words than on letters, it struggles significantly with noisy recordings, such as the example illustrated in Figure~\ref{fig_hayi_spectrogram_commented}.
A sample of transcriptions\footnotemark[1]{} \footnotetext[1]{https://osf.io/g83vs/?view\_only=19403ae222d44fe2bdd669c12fbce017}  in different scenarios is also provided, suggesting that direct multilingual ASR usage in classroom conditions may be less effective for short word- or letter-level inputs.

\begin{table}[ht]
\small
\centering
\caption{Transcription accuracy by question, using the MMS-1B model for speech-to-text.}
\label{tab_mms_18_results}
\begin{tabular}{lccc}
\hline
\textbf{Label} & \textbf{Samples} & \textbf{Correct} & \textbf{Accuracy (\%)} \\
\hline
d    & 503  & 28   & 5.57  \\
ewe  & 689  & 282  & 40.93 \\
hayi & 688  & 40   & 5.81  \\
hl   & 423  & 1    & 0.24  \\
v    & 796  & 4    & 0.50  \\
n    & 1139 & 8    & 0.70  \\
molo & 709  & 177  & 24.96 \\
kude & 557  & 64   & 11.49 \\
ng   & 538  & 2    & 0.37  \\
inja & 691  & 195  & 28.22 \\
\hline
\end{tabular}
\end{table}

\section{Conclusion}

This paper presents a cost-effective method for collecting and labeling child speech data in a low-resource language, and for training an AI model to assess reading proficiency, tailored for early-grade learners. We introduce and make available for the community a challenging dataset of Xhosa child speech~\cite{chevtchenko2025dataset}, recorded under realistic classroom conditions that include noise and multiple pronunciation attempts. The dataset contains 14,971 audio recordings, each containing a single word or letter, which are part of the Early Grade Reading Assessment (EGRA) system. The recordings were labeled online by three independent markers, and a sample was validated by an expert EGRA reviewer to confirm reliability. 

Three state-of-the-art speech recognition architectures were fine-tuned as binary classifiers and the experimental results indicate that data balancing and multi-question training strategies substantially affect performance. In particular, wav2vec 2.0 and Whisper are generally found to outperform HuBERT, achieving lower false positive and false negative rates. Although wav2vec 2.0 and Whisper demonstrate comparable accuracy, wav2vec 2.0 demands fewer computational resources during the training phase. Further comparisons with a large-scale multilingual ASR model (MMS-1B) confirm the effectiveness of the proposed approach, especially for short, noisy speech segments in a classroom setting. 

Future work is planned to focus on further improving the sample efficiency of the models, while providing a phoneme-based feedback to the user. Preliminary experiments on another LRL dataset, Sepedi, indicate similar performance for wav2vec 2.0 and this additional dataset is expected to be released in forthcoming research. Ultimately, the goal of this project is to develop a cost-effective reading tutoring system for early-grade learners, thereby supporting the preservation and educational advancement of low-resource languages.


%
%
%
\newpage
\bibliographystyle{splncs04}
\bibliography{cas-refs}

\begin{thebibliography}{10}
\providecommand{\url}[1]{\texttt{#1}}
\providecommand{\urlprefix}{URL }
\providecommand{\doi}[1]{https://doi.org/#1}

\bibitem{abaskohi2023persian}
Abaskohi, M., et~al.: Automatic speech recognition for speech assessment of persian preschool children. In: Proceedings of the International Conference on Speech and Language Processing. pp. 123--130 (2023)

\bibitem{bachiri2024integrating}
Bachiri, Y.A., Mouncif, H., Bouikhalene, B., Hamzaoui, R.: Integrating ai-based speech recognition technology to enhance reading assessments within morocco’s tarl program. Turkish Online Journal of Distance Education  \textbf{25}(4),  1--15 (2024)

\bibitem{baevski2020wav2vec}
Baevski, A., Zhou, Y., Mohamed, A., Auli, M.: wav2vec 2.0: A framework for self-supervised learning of speech representations. Advances in neural information processing systems  \textbf{33},  12449--12460 (2020)

\bibitem{bai2021automatic}
Bai, Y., Tejedor-Garc{\'i}a, C., Hubers, F., Cucchiarini, C., Strik, H.: Automatic speech recognition technology and reading skill development in primary school. In: ICERI2021 Proceedings. pp. 6188--6195. IATED (2021)

\bibitem{bai20b_interspeech}
Bai, Y., Hubers, F., Cucchiarini, C., Strik, H.: Asr-based evaluation and feedback for individualized reading practice. In: Interspeech 2020. pp. 3870--3874 (2020). \doi{10.21437/Interspeech.2020-2842}

\bibitem{chevtchenko2025dataset}
Chevtchenko, S., Navas, N., Vale, R., Ubaudi, F., Lucwaba, S., Ardington, C., Afshar, S., Antoniou, M., Afshar, S.: {EGRA}-{Xhosa}-14.9k: Annotated child reading audio dataset. \url{https://research-data.westernsydney.edu.au/published/7dfe822035f011f096a41d0408cdc7bb} (2025). \doi{https://doi.org/10.26183/93x0-qy45}

\bibitem{okwugbe2024fonigbo}
Dossou, B.F.P., Emezue, O.: Okwugbé: End-to-end speech recognition for fon and igbo. In: Proceedings of the 2024 International Conference on Speech and Language Technology. pp. 123--134 (2021)

\bibitem{doumbouya2021radio}
Doumbouya, M.L., et~al.: Using radio archives for low-resource speech recognition: Towards an intelligent virtual assistant for illiterate users. In: Proceedings of the IEEE International Conference on Acoustics, Speech and Signal Processing (ICASSP). pp. 1234--1238 (2021)

\bibitem{Dubeck2015}
Dubeck, M.M., Gove, A.: The early grade reading assessment ({EGRA)}: Its theoretical foundation, purpose, and limitations. Int. J. Educ. Dev.  \textbf{40},  315--322 (Jan 2015)

\bibitem{henkel2025literacy}
Henkel, O., Horne-Robinson, H., Hills, L., Roberts, B., McGrane, J.: Supporting literacy assessment in west africa: Using state-of-the-art speech models to assess oral reading fluency. International Journal of Artificial Intelligence in Education pp. 1--22 (2025)

\bibitem{hsu2021hubert}
Hsu, W.N., Bolte, B., Tsai, Y.H.H., Lakhotia, K., Salakhutdinov, R., Mohamed, A.: Hubert: Self-supervised speech representation learning by masked prediction of hidden units. IEEE/ACM transactions on audio, speech, and language processing  \textbf{29},  3451--3460 (2021)

\bibitem{jain2023child}
Jain, R., Barcovschi, A., Yiwere, M.Y., Bigioi, D., Corcoran, P., Cucu: A wav2vec2-based experimental study on self-supervised learning methods to improve child speech recognition. IEEE Access  \textbf{11},  46938--46948 (2023)

\bibitem{kim2022}
Kim, S., Park, S., Kim, K., Jung, K., So, S., Kim, M.: Automatic pronunciation assessment using self-supervised speech representation learning. arXiv preprint arXiv:2204.03863  (2022)

\bibitem{magueresse2020low}
Magueresse, A., Carles, V., Heetderks, E.: Low-resource languages: A review of past work and future challenges. arXiv preprint arXiv:2006.07264  (2020)

\bibitem{mesham2021low}
Mesham, S., Hayward, L., Shapiro, J., Buys, J.: Low-resource language modelling of south african languages. arXiv preprint arXiv:2104.00772  (2021)

\bibitem{Namasivayam2020}
Namasivayam, A.K., Coleman, D., O'Dwyer, A., van Lieshout, P.: Speech sound disorders in children: An articulatory phonology perspective. Front. Psychol.  \textbf{10} (Jan 2020)

\bibitem{obiang2024yoruba}
Obiang, J.P., et~al.: Improving tone recognition performance using wav2vec 2.0-based learned representation in yoruba, a low-resourced language. Journal of Speech and Language Technology  \textbf{37},  101--120 (2024)

\bibitem{peng2023md}
Peng, L., Gao, Y., Bao, R., Li, Y., Zhang, J.: End-to-end mispronunciation detection and diagnosis using transfer learning. Applied Sciences  \textbf{13}(11), ~6793 (2023)

\bibitem{pratap2023mms}
Pratap, V., Tjandra, A., Shi, B., Tomasello, P., Babu, A., Kundu, S., Elkahky, A., Ni, Z., Vyas, A., Fazel-Zarandi, M., Baevski, A., Adi, Y., Zhang, X., Hsu, W.N., Conneau, A., Auli, M.: Scaling speech technology to 1,000+ languages. arXiv  (2023)

\bibitem{radford2023robust}
Radford, A., Kim, J.W., Xu, T., Brockman, G., McLeavey, C., Sutskever, I.: Robust speech recognition via large-scale weak supervision. In: International conference on machine learning. pp. 28492--28518. PMLR (2023)

\bibitem{ranathunga2023low}
Ranathunga, S., Lee, E.S.A., Prifti~Skenduli, M., Shekhar, R., Alam, M., Kaur, R.: Neural machine translation for low-resource languages: A survey. ACM Computing Surveys  \textbf{55}(11),  1--37 (2023)

\bibitem{sabu2018automatic}
Sabu, K., Rao, P.: Automatic assessment of children’s oral reading using speech recognition and prosody modeling. CSI Transactions on ICT  \textbf{6},  221--225 (2018)

\bibitem{shekar2023gop}
Shekar, J., et~al.: Assessment of non-native speech intelligibility using wav2vec2-based mispronunciation detection and multi-level goodness of pronunciation transformer. Journal of Speech Processing  \textbf{34},  123--145 (2023)

\bibitem{shivakumar2020transfer}
Shivakumar, P.G., Georgiou, P.: Transfer learning from adult to children for speech recognition: Evaluation, analysis and recommendations. Computer speech \& language  \textbf{63},  101077 (2020)

\bibitem{Sobti2024}
Sobti, R., Guleria, K., Kadyan, V.: Comprehensive literature review on children automatic speech recognition system, acoustic linguistic mismatch approaches and challenges. Multimed. Tools Appl.  (Mar 2024)

\bibitem{yildiz2025can}
Y{\i}ld{\i}z, M., Keskin, H.K., Oyucu, S., Hartman, D.K., Temur, M., Aydo{\u{g}}mu{\c{s}}, M.: Can artificial intelligence identify reading fluency and level? comparison of human and machine performance. Reading \& Writing Quarterly  \textbf{41}(1),  66--83 (2025)

\end{thebibliography}
%




\end{document}